\documentclass[fleqn,10pt]{wlscirep}
\usepackage[utf8]{inputenc}
\usepackage{multirow}
\usepackage[T1]{fontenc}

\title{AI-Driven CT-based quantification, staging and short-term outcome prediction of COVID-19 pneumonia}

\author[*,1,2,3]{Guillaume Chassagnon, MD PhD}
\author[4,5,6,7]{Maria Vakalopoulou, PhD\footnote{Dr. Guillaume Chassagnon \& Dr. Maria Vakalopoulou have equally contributed to this work}}
\author[4,6,7]{Enzo Battistella, MsC}
\author[8,9]{Stergios Christodoulidis, PhD}
\author[1]{Trieu-Nghi Hoang-Thi, MD}
\author[1]{Severine Dangeard, MD}
\author[6,7]{Eric Deutsch, MD PhD}
\author[8,9]{Fabrice Andre, MD PhD}
\author[1]{Enora Guillo, MD}
\author[1]{Nara Halm, MD}
\author[1]{Stefany El Hajj, MD}
\author[1]{Florian Bompard, MD}
\author[1]{Sophie Neveu, MD}
\author[1]{Chahinez Hani, MD}
\author[1]{Ines Saab, MD}
\author[1]{Ali\'enor Campredon, MD}
\author[1]{Hasmik Koulakian, MD}
\author[1]{Souhail Bennani, MD}
\author[1]{Gael Freche, MD}
\author[15]{Aurelien Lombard, MsC}
\author[2,10]{Laure Fournier, MD PhD}
\author[10]{Hippolyte Monnier, MD}
\author[10]{T\'eodor Grand, MD}
\author[2,11]{Jules Gregory, MD}
\author[2,12]{Antoine Khalil, MD PhD}
\author[2,12]{Elyas Mahdjoub, MD}
\author[13]{Pierre-Yves Brillet, MD PhD}
\author[13]{St\'ephane Tran Ba, MD}
\author[2,14]{Val\'erie Bousson, MD PhD}
\author[1,2,3]{Marie-Pierre Revel, MD PhD}
\author[4,7,15]{Nikos Paragios, PhD\footnote{Corresponding author: n.paragios@therapanacea.eu}}
\affil[1]{Radiology Department, Hopital Cochin – AP-HP.Centre Universit\'e de Paris, 27 Rue du Faubourg Saint-Jacques, 75014 Paris, France}
\affil[2]{Universit\'e de Paris, 85 boulevard Saint-Germain, 75006 Paris, France}
\affil[3]{Inserm U1016, Institut Cochin, 22 rue M\'echain, 75014 Paris, France}
\affil[4]{Universit\'e Paris-Saclay, CentraleSup\'elec, Math\'ematiques et Informatique pour la Complexit\'e et les Syst\'emes, Gif-sur-Yvette, France, 3 Rue Joliot Curie, 91190 Gif-sur-Yvette, France}
\affil[5]{Universit\'e Paris-Saclay, CentraleSup\'elec, Inria, Gif-sur-Yvette, France}
\affil[6]{Universit\'e Paris-Saclay, Institut Gustave Roussy, Inserm 981 Molecular Radiotherapy and Innovative Therapeutics, 114 Rue Edouard Vaillant, 94800 Villejuif, France}
\affil[7]{Gustave Roussy-CentraleSupélec-TheraPanacea, Noesia Center of Artificial Intelligence in Radiation Therapy and Oncology, Gustave Roussy Cancer Campus, Villejuif, France}
\affil[8]{Universit\'e Paris-Saclay, Institut Gustave Roussy, Inserm 1030 Predictive Biomarkers and New Therapeutic Strategies in Oncology, 114 Rue Edouard Vaillant, 94800 Villejuif, France}
\affil[9]{Universit\'e Paris-Saclay, Institut Gustave Roussy, Prism Precision Medicine Center, 114 Rue Edouard Vaillant, 94800 Villejuif, France}
\affil[10]{Radiology Department, Hopital Europ\'een Georges Pompidou – AP-HP.Centre Universit\'e de Paris, 20 Rue Universit\'e Paris-Saclay, 75015 Paris, France}
\affil[11]{Radiology Department, Hopital Beaujon – AP-HP.Nord Universit\'e de Paris, 100 Boulevard du G\'en\'eral Leclerc, 92110 Clichy}
\affil[12]{Radiology Department, Hopital Bichat – AP-HP.Nord Universit\'e de Paris, 46 Rue Henri Huchard, 75018 Paris, France}
\affil[13]{Radiology Department, Hopital Avicenne – AP-HP.Hopitaux universitaires Paris Seine-Saint-Denis, 125 Rue de Stalingrad, 93000 Bobigny, France}
\affil[14]{Radiology Department, Hopital Lariboisi\'ere – AP-HP.Nord Universit\'e de Paris, 2 Rue Ambroise Par\'e, 75010 Paris, France}
\affil[15]{TheraPanacea, 27 Rue du Faubourg Saint-Jacques, 75014 Paris, France}


\begin{abstract}
Chest computed tomography (CT) is widely used for the management of Coronavirus disease 2019 (COVID-19) pneumonia because of its availability and rapidity~\cite{zu2020coronavirus,bai2020performance,bernheim2020chest} . The standard of reference for confirming COVID-19 relies on microbiological tests but these tests might not be available in an emergency setting and their results are not immediately available, contrary to CT. In addition to its role for early diagnosis, CT has a prognostic role by allowing visually evaluating the extent of COVID-19 lung abnormalities~\cite{li2020ct,yuan2020association}. The objective of this study is to address prediction of short-term outcomes, especially need for mechanical ventilation. In this multi-centric study, we propose an end-to-end artificial intelligence solution for automatic quantification and prognosis assessment by combining automatic CT delineation of lung disease meeting expert’s performance and data-driven identification of biomarkers for its prognosis. AI-driven combination of variables with CT-based biomarkers offers perspectives for optimal patient management given the shortage of intensive care beds and ventilators~\cite{truog2020toughest,white2020framework}.
\end{abstract}

\begin{document}

\flushbottom
\maketitle
%
%
\thispagestyle{empty}


\begin{figure}[b!]
\centering
\includegraphics[width=\linewidth]{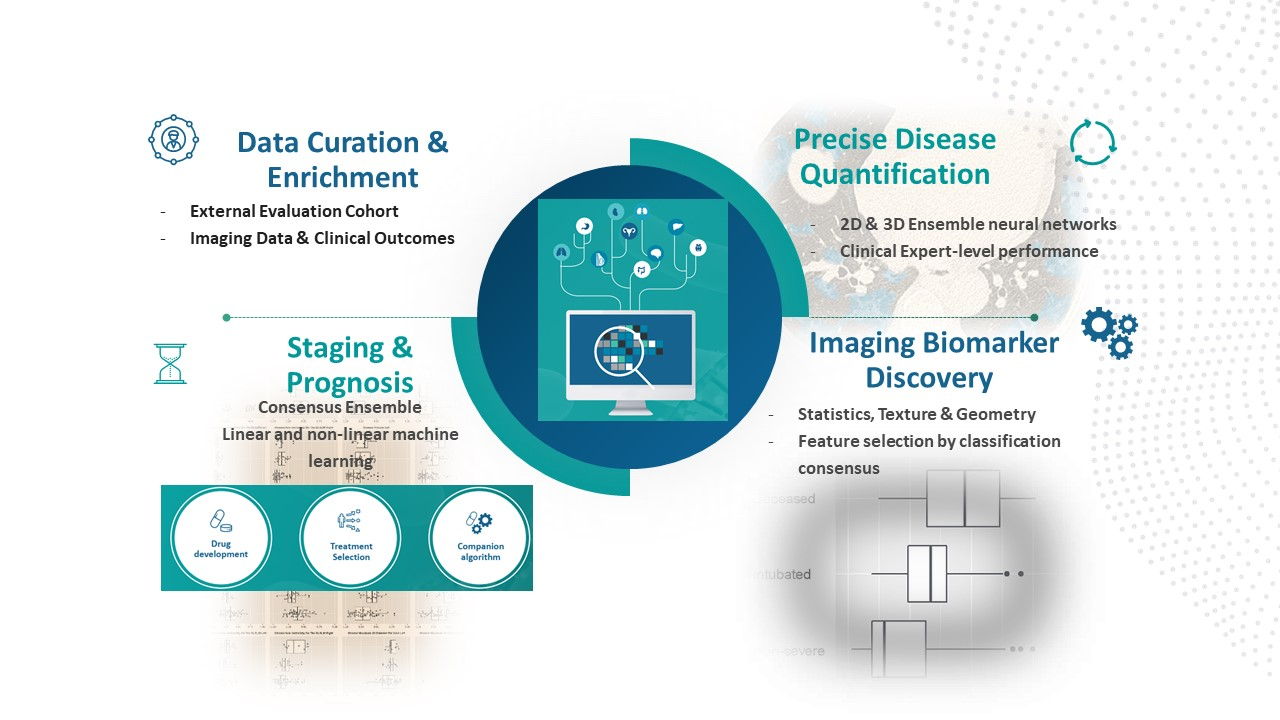}
\caption{Overview of the method for CT-based quantification, staging and prognosis of COVID-19. (i) Two independent cohorts with quantification based on ensemble 2D \& 3D consensus neural networks reaching expert-level annotations on massive evaluation, (ii) Consensus-driven bio(imaging)-marker selection on the principle of prevalence across methods leading to variables highly-correlated with outcomes \& (iii) Consensus of linear \& non-linear classification methods for staging and prognosis reaching optimal performance (minimum discrepancy between training \& testing).}
\label{fig:fig1}
\end{figure}

\section*{Main}
COVID-19 has emerged in December 2019 in the city of Wuhan in China~\cite{zhu2020novel} and disseminated around the world, leading the World Health Organization to declare the COVID-19 outbreak a pandemic. The disease is caused by the SARS-Cov-2 virus and the leading cause of death is respiratory failure due to severe viral pneumonia~\cite{zhou2020clinical}. Chest computed tomography (CT) has rapidly gained a major role for COVID-19 diagnosis. Indeed, despite being considered as the gold standard to make a definitive diagnosis, reverse transcription polymerase chain reaction (RT-PCR) suffers from false negatives, shortage of available supply test kits and long turnaround times~\cite{ai2020correlation,fang2020sensitivity,xie2020chest}. 

Artificial intelligence has gained significant attention during the past decade and many applications have been proposed in medical imaging, including segmentation and characterization tasks such as lung cancer screening on CT~\cite{chassagnon2020artificial,litjens2017survey,ardila2019end}. A few studies have already reported deep learning to diagnose COVID-19 pneumonia on chest radiograph~\cite{wang2020covid} or CT~\cite{li2020artificial} . Other authors used deep-learning to quantify COVID-19 disease extent on CT but none of them used a multi-centric cohort while providing comparisons with segmentations done by radiologists~\cite{chaganti2020quantification,huang2020serial}. Disease extent is the only parameter that can be visually estimated on chest CT to quantify disease severity~\cite{li2020ct,yuan2020association}, but visual quantification is difficult and usually coarse. Several AI-based tools have been recently developed to quantify interstitial lung diseases (ILD)~\cite{jacob2016mortality,humphries2017idiopathic,10.1007/978-3-030-00937-3_75,anthimopoulos2018semantic}, which share common CT features with COVID-19 pneumonia, especially a predominance of ground glass opacities. In this study, we investigated a fully automatic method (Figure~\ref{fig:fig1}) for disease quantification, staging and short-term prognosis. The approach relied on (i) a disease quantification solution that exploited 2D \& 3D convolutional neural networks using an ensemble method, (ii) a biomarker discovery approach sought to determine the share space of features that are the most informative for staging \& prognosis, \& (iii) an ensemble robust supervised classification method to distinguish patients with severe vs non-severe short-term outcome and among severe patients those intubated and those who did not survive.

\begin{figure}[t!]
\centering
\includegraphics[width=\linewidth]{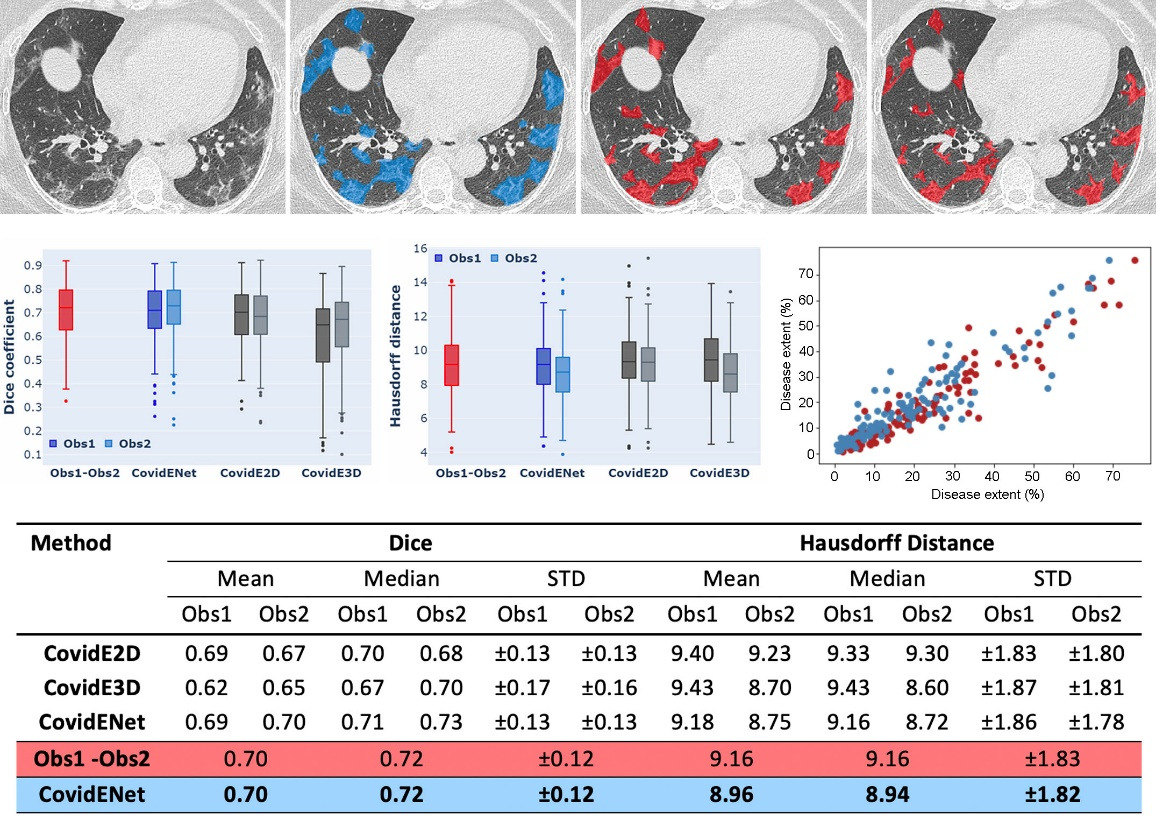}
\caption{Comparison between automated and manual segmentations. Delineation of the diseased areas on chest CT in a COVID-19 patient: First Row:  input, AI-segmentation, expert I-segmentation, expert II-segmentation. Second Row: Box-Plot Comparisons in terms of Dice similarity and Haussdorf between AI-solution, expert I \& expert II, \& Plot of correlation between disease extent automatically measured and the average disease extent measured from the $2$ manual segmentations. Disease extent is expressed as the percentage of lung affected by the disease. Third row: statistical measures on comparisons between AI, expert I, and expert II segmentations.}
\label{fig:fig2}
\end{figure}

\section*{Part I: Disease Quantification}
In the context of this work, we report a deep learning-based segmentation tool to quantify COVID-19 disease and lung volume. For this purpose, we used an ensemble network approach inspired by the AtlasNet framework~\cite{10.1007/978-3-030-00937-3_75}. We investigated a combination of 2D slice-based~\cite{badrinarayanan2017segnet} and 3D patch-based ensemble architectures~\cite{cciccek20163d}. The development of the deep learning-based segmentation solution was done on the basis of a multi-centric cohort of $478$ unenhanced chest CT scans (208,668 slices) of COVID-19 patients with positive RT-PCR. The multicentric dataset was acquired at $6$ Hospitals, equipped with $4$ different CT models from $3$ different $91$ manufacturers, with different acquisition protocols and radiation dose (Table~\ref{tab:tab1}). Fifty CT exams from $3$ centers were used for training and $130$ CT exams from $3$ other centers were used for test (Table~\ref{tab:tab2}). Disease and lung were delineated on all $23,423$ images used as training dataset, and on only $20$ images per exam but by $2$ independent annotators in the test dataset ($2,600$ images). The overall annotation effort took approximately $800$ hours and involved $15$ radiologists with $1$ to $7$ years of experience in chest imaging. The consensus between manual ($2$ annotators) and automated segmentation was measured using the Dice similarity score (DSC)~\cite{dice1945measures} and the Haussdorf distance (HD). The CovidENet performed equally well to trained radiologists in terms of DSCs and better in terms HD (Figure~\ref{fig:fig2}). The mean/median DSCs between the two expert’s annotations on the test dataset were $0.70/0.72$ for disease segmentation. For the same task, DSCs between CovidENet and the manual segmentations were $0.69/0.71$ and $0.70/0.73$. In terms of HDs, the observed average value between the two experts was $9.16$mm while it was $8.96$mm between CovidENet and the two experts. When looking at disease extent, defined as the percentage of lung affected by the disease, we found no significant difference between automated segmentation and the average of the two manual segmentations ($19.9\%$ $\pm17.7$ [$0.5 - 73.2$] vs $19.5\%$ $\pm16.5$ [$1.1 - 75.7$]; p= $0.352$).

\begin{figure}[t]
\centering
  \includegraphics[width=3.25in]{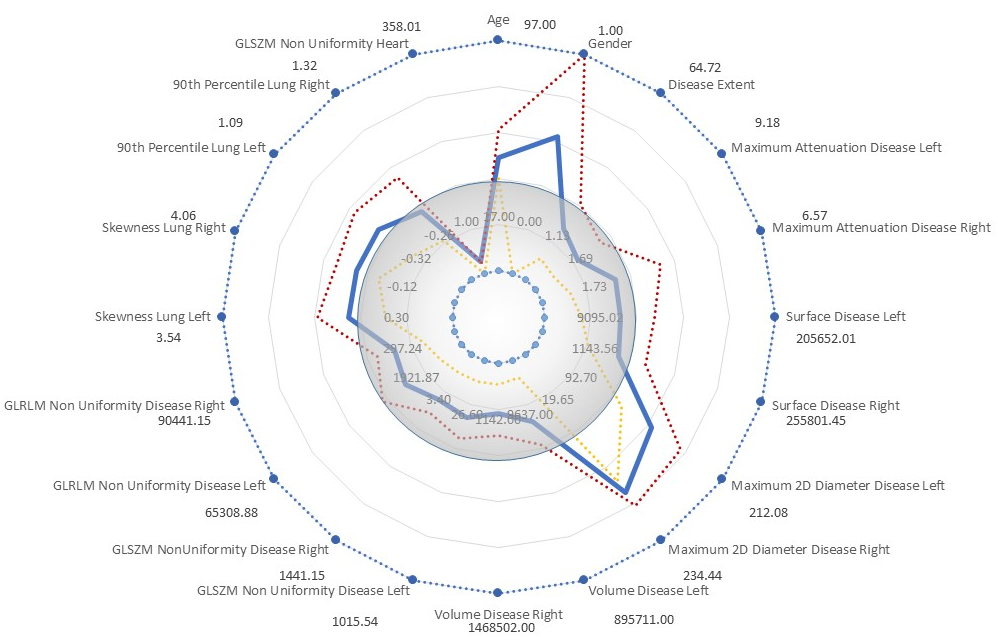} \\
  \includegraphics[width=3.25in]{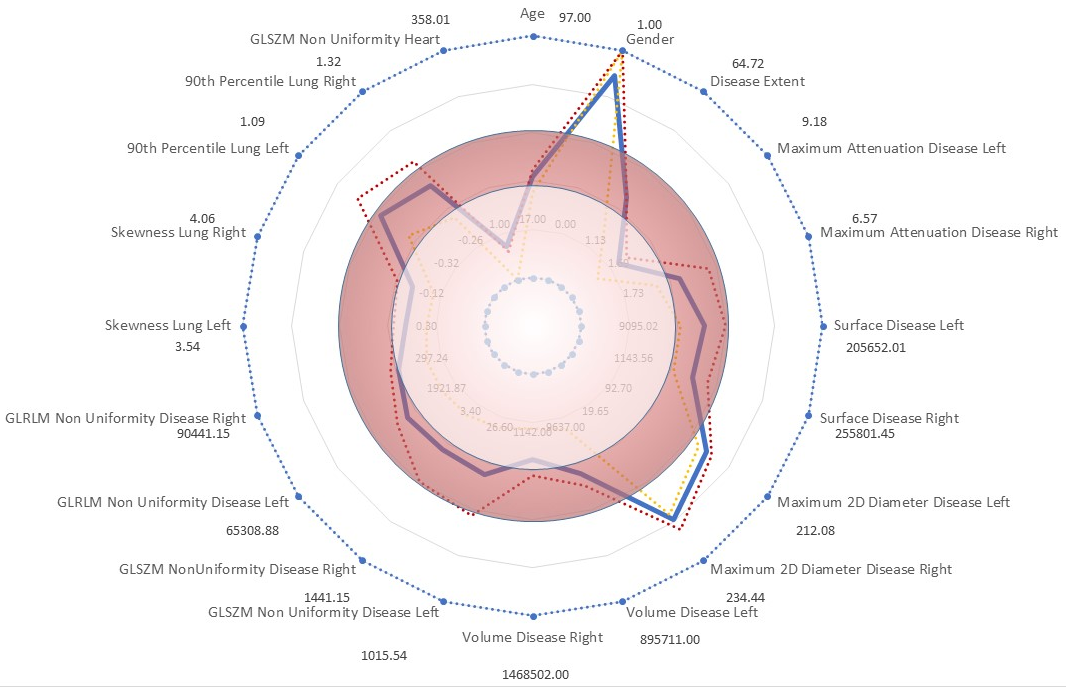}
    \includegraphics[width=3.25in]{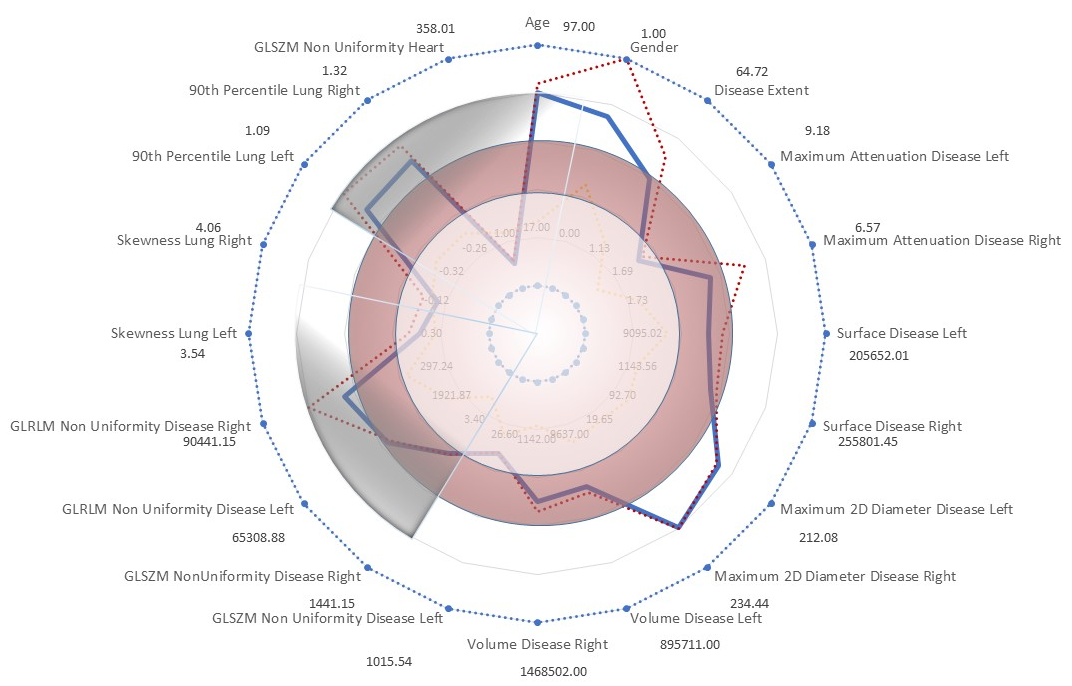}
\caption{Spider-chart distribution of features depicting their minimum and maximum values [mean value (blue), 70\% percentile (yellow) and 90\% percentile (red) lines] with respect to the different outcomes with the following order: top: non-severe, bottom left: intensive care support \& bottom right: deceased in the testing set. White and red circles represent respectively 40\% and 60\% of the maximum value of each feature. Clear separation was observed on these feature space with respect to the non-severe \& severe cases. In terms of deceased versus intensive care patients, notable difference were observed with respect to three variables, the age of the patient, the condition of the healthy lung and the non-uniformity of the disease (indicated with gray in the spider-chart).}
\label{fig:fig3b}
\end{figure}

\section*{Part II: Imaging Biomarker Discovery}
To assess the prognostic value of the Chest computed tomography (CT) an extended multi-centric data set was built. We reviewed outcomes in patient charts within the $4$ days following chest CT and divided the patients in $3$ groups: those who didn’t survive, those who required mechanical ventilation and those who were still alive and not intubated. Out of the $478$ included patients, $27$ died ($6\%$) and $83$ were intubated ($17\%$), forming a group of $110$ patients with severe short-term outcome ($23\%$). Data of $383$ patients from $3$ centers were used for training and those of $85$ patients from $3$ other centers composed an independent test dataset (Table~\ref{tab:tab3}). Radiomics-based prognosis gained significant attention in the recent years towards predicting treatment outcomes~\cite{sun2018radiomics}. In this study we have adopted a similar strategy, we extracted $107$ features related to first order, higher order statistics, texture and shape information for lungs, disease extent and heart. Feature selection was performed on a basis of predictive value consensus. We created several representative partitions
$117$ of the training set ($80\%$ training and $20\%$ validation) and run $13$ different supervised classification methods towards optimal separation of the observed clinical ground truth between severe and non-severe cases (Table~\ref{tab:tab4}). The features that were shared between the different classifiers were retained as robust imaging biomarkers using a cut-off probability of $0.25$ and were aggregated to patients’ age and gender (Table~\ref{tab:tab5}). In total $12$ features were retained for the prognosis part and included age, gender, disease extent, descriptors of disease heterogeneity and extension, features of healthy lung and a descriptor of cardiac heterogeneity. Correlations for some these features and the clinical outcome are presented in Figure~\ref{fig:fig3} while a representation of these feature space with respect to the different classes is presented in Figure~\ref{fig:fig3b}.

\section*{Part III: Staging and Prognosis}
\begin{figure}[t!]
\centering
\includegraphics[width=\linewidth]{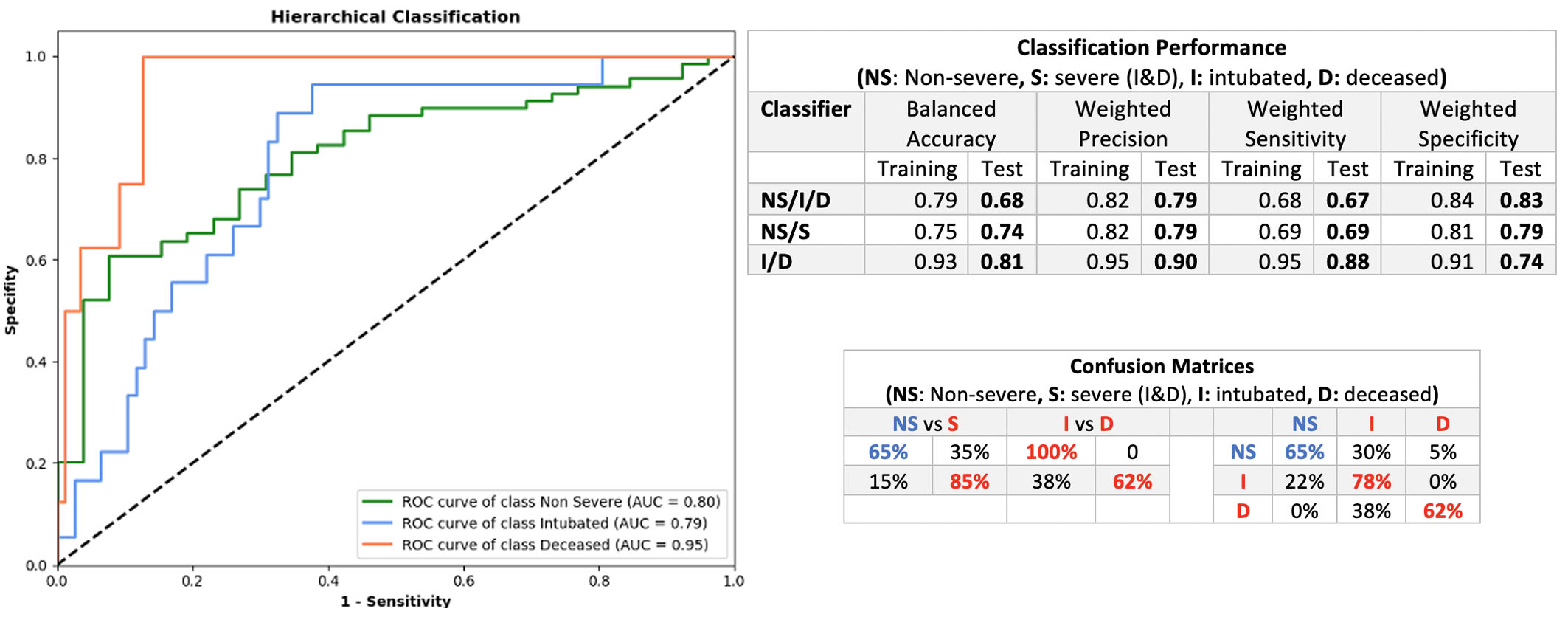}
\caption{Classification performance of dual \& aggregated classifiers with respect to the non-severe vs severe case, the intubated vs deceased case and the three classes. Sensitivity and confusion tables are presented with respect to the different classification problems.}
\label{fig:fig4}
\end{figure}

\begin{figure}[t!]
\centering
\includegraphics[width=\linewidth]{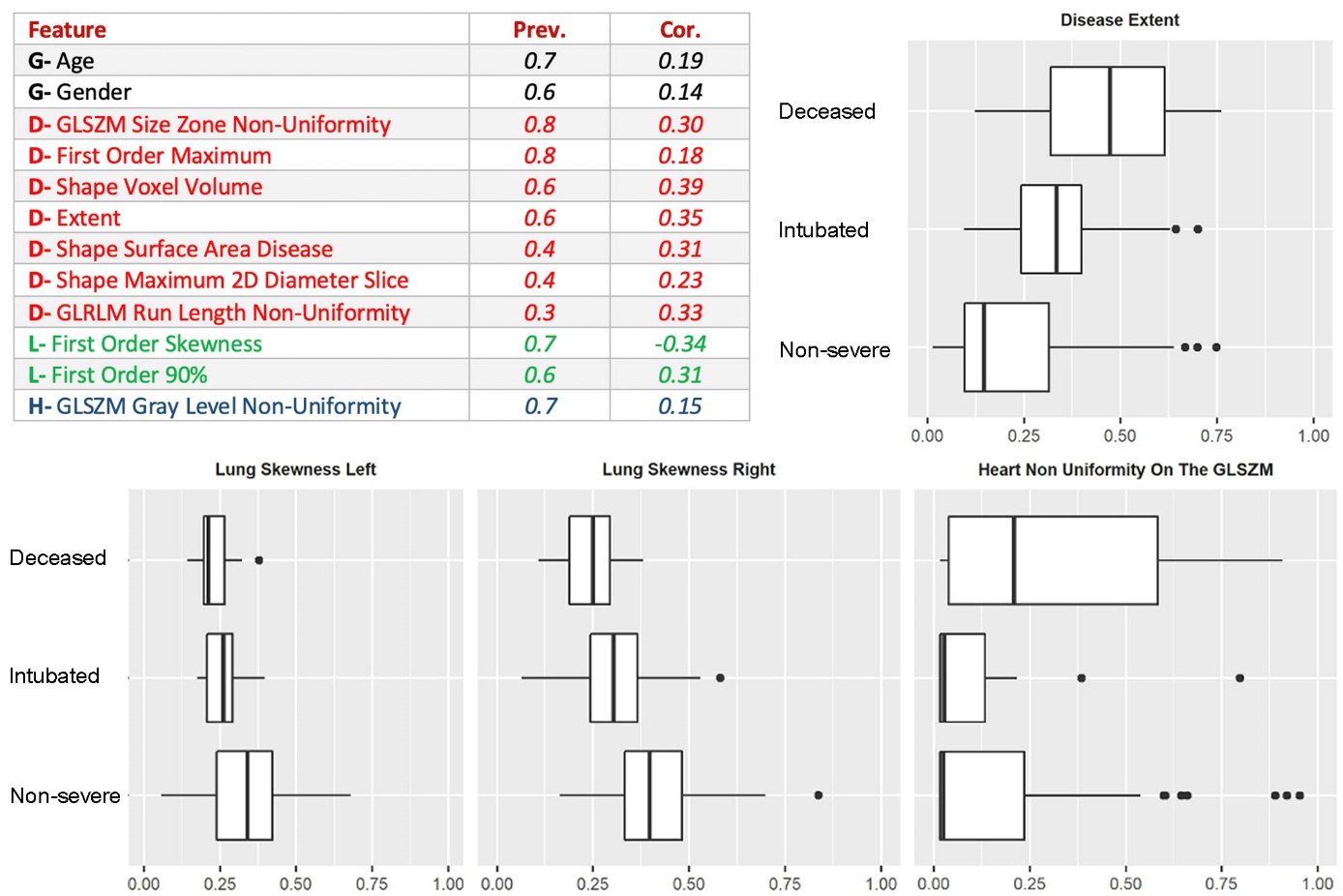}
\caption{Discovery of – imaging-biomarkers through consensus. Generic variables (G: age, sex), disease related variables (D: extent, volume, maximum diameter, etc.), lung variables (L: skewness, etc.) as well as heart related variables (H: non-uniformity) have been automatically selected. The prevalence of the features as well as their distribution with respect to the different classes is presented for some of them, with rather clear separation and strong correlations with ground truth.}
\label{fig:fig3}
\end{figure}

The staging/prognosis was implemented using a hierarchical classification principle, targeting first staging and subsequently prognosis. The staging component sought to separate patients with severe and non-severe short-term outcomes, while the prognosis sought to predict the risk of decease among severe patients. On the basis of the feature selection step, the machine learning algorithms that had a balanced accuracy greater than $60\%$ on validation were considered. The selection of these methods was done on the basis of minimum discrepancy between performance on training and internal validation sub-training data set. We have built two sequential classifiers using this ensemble method, one to determine the severe cases and a second to predict survival. The classifier aiming to separate patients with severe and non-severe short-term outcomes had a balanced accuracy of $74\%$, a weighted precision of $79\%$, a weighted sensitivity of $69\%$ and specificity of $79\%$ to predict a severe short-term outcome (Figure~\ref{fig:fig4}, Table~\ref{tab:tab6}). The performance of the second classifier aiming to differentiate between intubated and deceased patients was even higher with a balanced accuracy of $81\%$ (Figure~\ref{fig:fig4}, Table~\ref{tab:tab7}). The hierarchical classifiers combing the $3$ classes had a balanced accuracy of $68\%$, a weighted precision of $79\%$, a weighted sensitivity of $67\%$ and specificity of $83\%$ (Figure~\ref{fig:fig4}). It was observed that prognosis performance difference between training and external cohort testing was low, suggesting that the most important information present at CT scans was recovered, and additional information should be integrated in order to fully explain the outcome.

\section*{Part IV: Conclusions}
In conclusion, artificial intelligence enhanced the value of chest CT by providing fast accurate, and precise disease extent quantification and by helping to identify patients with severe short-term outcomes. This could be of great help in the current context of the pandemic with healthcare resources under extreme pressure. In a context where the sensitivity of RT-PCR has been shown to be low, such as $63\%$ when perform on nasal swab~\cite{wang2020detection}, chest CT has been shown to provide higher sensitivity for diagnosis of COVID-19 as compared with initial RT-PCR from pharyngeal swab samples~\cite{ai2020correlation}. The current COVID-19 pandemic requires implementation of rapid clinical triage in healthcare facilities to categorize patients into different urgency categories~\cite{CDC}, often occurring in the context of limited access to biological tests. Beyond the diagnostic value of CT for COVID-19, our study suggests that AI should be part of the triage process. The developed tool will be made publicly available. Our prognosis and staging method achieved state of the art results through the deployment of a highly robust ensemble classification strategy with automatic feature selection of imaging biomarkers and patients’ characteristics available within the image’ metadata. In terms of future work, the continuous enrichment of the data base with new examples is a necessary action on top of updating the outcome of patients included in the study. The integration of non-imaging data and other related clinical and categorical variables such as lymphopenia, the D-dimer level and other comorbidities~\cite{zhou2020clinical,tang2020abnormal,onder2020case,guo2020diabetes} is a necessity towards better understanding the disease and predicting the outcomes. This is clearly demonstrated from the inability of any of the state-of-the art classification methods (including neural networks and multi-layer perceptron models) to predict the outcome with a balanced accuracy greater to $80\%$ on the training data. Our findings could have a strong impact in terms of (i) patient stratification with respect to the different therapeutic strategies, (ii) accelerated drug development through rapid, reproducible and quantified assessment of treatment response through the different mid/end-points of the trial, and (iii) continuous monitoring of patient’s response to treatment.

\section*{Methods}
\subsection*{Study Design and Participants}
This retrospective multi-center study was approved by our Institutional Review Board (AAA-2020-08007) which waived the need for patients’ consent. Patients diagnosed with COVID-19 from March $4$th to $29$th at six large University Hospitals were eligible if they had positive PCR-RT and signs of COVID-19 pneumonia on unenhanced chest CT. A total of $478$ patients formed the full dataset ($208,668$ CT slices). Only one CT examination was included for each patient. Exclusion criteria were (i) contrast medium injection and (ii) important motion artifacts.

For the COVID-19 radiological pattern segmentation part, $50$ patients from $3$ centers (A: $20$ patients; B: $15$ patients, C: $15$ patients) were included to compose a training and validation dataset, $130$ patients from the remaining $3$ centers (D: $50$ patients; E: $50$ patients, F: $30$ patients) were included to compose the test dataset (Table~\ref{tab:tab2}). The proportion between the CT manufacturers in the datasets was pre-determined in order to maximize the model generalizability while taking into account the data distribution.

For the radiomics driven prognosis study, $298$ additional patients from centers A ($96$ patients), B ($64$ patients) and D ($138$ patients) were included to increase the size of the dataset. Data of $383$ patients from $3$ centers (A, B and D) were used for training and those of 85 patients from 3 other centers (C, E, F) composed an independent test set (Table~\ref{tab:tab3}). Only one CT examination was included for each patient. Exclusion criteria were (i) contrast medium injection and (ii) important motion artifacts. For short-term outcome assessment, patients were divided into $2$ groups: those who died or were intubated in the $4$ days following the CT scan composed the severe short-term outcome subgroup, while the others composed the non-severe short-term outcome subgroup.

\subsection*{CT Acquisitions}
Chest CT exams were acquired on $4$ different CT models from $3$ manufacturers (Aquilion Prime from Canon Medical Systems, Otawara, Japan; Revolution HD from GE Healthcare, Milwaukee, WI; Somatom Edge and Somatom AS+ from Siemens Healthineer, Erlangen, Germany). The different acquisition and reconstruction parameters are summarized in Table~\ref{tab:tab1}. CT exams were mostly acquired at $120$ (n=$103/180$; $57\%$) and $100$ kVp (n=$76/180$; $42\%$). Images were reconstructed using iterative reconstruction with a $512\times512$ matrix and a slice thickness of $0.625$ or $1$ mm depending on the CT equipment. Only the lung images reconstructed with high frequency kernels were used for analysis. For each CT examination, dose length product (DLP) and volume Computed Tomography Dose Index (CTDIvol) were collected.

\subsection*{Data Annotation}
Fifteen radiologists (GC, TNHT, SD, EG, NH, SEH, FB, SN, CH, IS, HK, SB, AC, GF and MB) with $1$ to $7$ years of experience in chest imaging participated in the data annotation which was conducted over a $2$-week period.
For the training and validation set for the COVID-19 radiological pattern segmentation, the whole CT examinations were manually annotated slice by slice using the open source software ITKsnap~\footnote{\url{http://www.itksnap.org}}. On each of the $23,423$ axial slices composing this dataset, all the COVID-19 related CT abnormalities (ground glass opacities, band consolidations, and reticulations) were segmented as a single class. Additionally, the whole lung was segmented to create another class (lung). To facilitate the collection of the ground truth for the lung anatomy, a preliminary lung segmentation was performed with Myrian XP-Lung software (version $1.19.1$, Intrasense, Montpellier, France) and then manually corrected.

As far as test cohort for the segmentation is concerned, $20$ CT slices equally spaced from the superior border of aortic arch to the lowest diaphragmatic dome were selected to compose a $2,600$ images dataset. Each of these images were systematically annotated by $2$ out of the $15$ participating radiologists who independently performed the annotation. Annotation consisted of manual delineation of the disease and manual segmentation of the lung without using any preliminary lung segmentation.

\subsection*{Deep Learning Construction}
The segmentation tool was built under the paradigm of ensemble methods using a 2D fully convolutional network together with the AtlasNet framework~\cite{10.1007/978-3-030-00937-3_75} and a 3D fully convolutional network~\cite{cciccek20163d}. The AtlasNet framework combines a registration stage of the CT scans to a number of anatomical templates and consequently utilizes multiple deep learning-based classifiers trained for each template. At the end, the prediction of each model is - to the original anatomy and a majority voting scheme is used to produce the final projection, combining the results of the different networks. A major advantage of the AtlasNet framework is that it incorporates a natural data augmentation by registering each CT scan to several templates. Moreover, the framework is agnostic to the segmentation model that will be utilized. For the registration of the CT scans to the templates, an elastic registration framework based on Markov Random Fields was used, providing the optimal displacements for each template~\cite{ferrante2017deformable}.

The architecture of the implemented segmentation models was based on already established fully convolutional neural network designs from the literature~\cite{badrinarayanan2017segnet,cciccek20163d}. Fully convolutional networks following an encoder decoder architecture both in 2D and 3D were developed and evaluated. For the 2D models the CT scans were separated on the axial view. The network included 5 convolutional blocks, each one containing two Conv-BN-ReLU layer successions. Maxpooling layers were also distributed at the end of each convolutional block for the encoding part. Transposed convolutions were used on the decoding part to restore the spatial resolution of the slices together with the same successions of layers. For the 3D pipeline, the model similarly consisted of five blocks with a down-sampling operation applied every two consequent Conv3D-BN-ReLU layers. Additionally, five decoding blocks were utilized for the decoding path, at each block a transpose convolution was performed in order to up-sample the input. Skip connections were also employed between the encoding and decoding paths. In order to train this model, cubic patches of size $64\times64\times64$ were randomly extracted within a close range of the ground truth annotation border in a random fashion. Corresponding cubic patches were also extracted from the ground truth annotation masks and the lung anatomy segmentation masks. To this end, we trained the model with the CT scan patch as input, the annotation patch as target and the lung anatomy annotation patch as a mask for calculating the loss function only within the lung region. In order to train all the models, each CT scan was normalized by cropping the Hounsfield units in the range [$-1024$, $1000$]. 

Regarding implementation details, 6 templates were used for the AtlasNet framework together with normalized cross correlation and mutual information as similarities metrics. The networks were trained using weighted cross entropy loss using weights depending on the appearance of each class and dice loss. Moreover, the 3D network was trained using a dice loss.
The Dice loss (DL) and weighted cross entropy (WCE) are defined as follows,

\begin{equation*}
DL = 1 - \dfrac{2pg+1}{p+g+1}, \hspace{5mm} WCE = -(\beta g \log(p) + (1-g) log(1-p))
\end{equation*}    
where $p$ is the predicted from the network value and $g$ the target/ ground truth value. $\beta$ is the weight given for the less representative class. For network optimization, we used only the class for the diseased regions.

\begin{figure}[t!]
\centering
\includegraphics[width=0.9\linewidth]{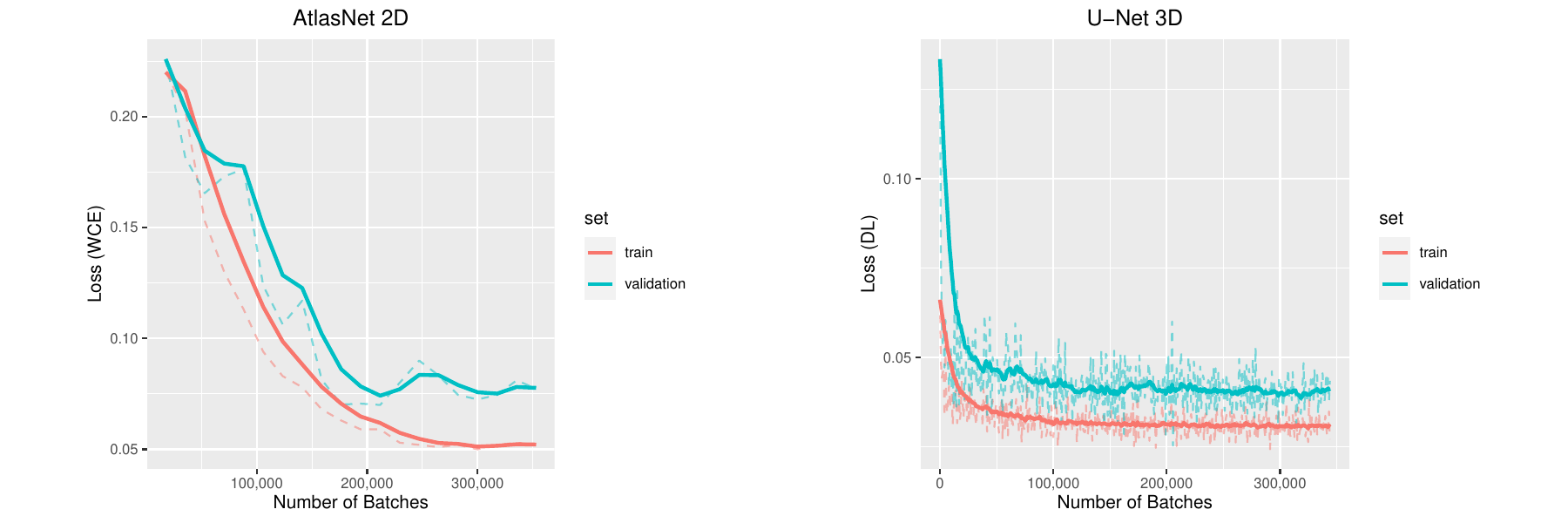}
\caption{Training and validation curves for one template of AtlasNet and the 3D U-Net.}
\label{fig:fig5}
\end{figure}

\begin{figure}[t!]
\centering
\includegraphics[width=0.9\linewidth]{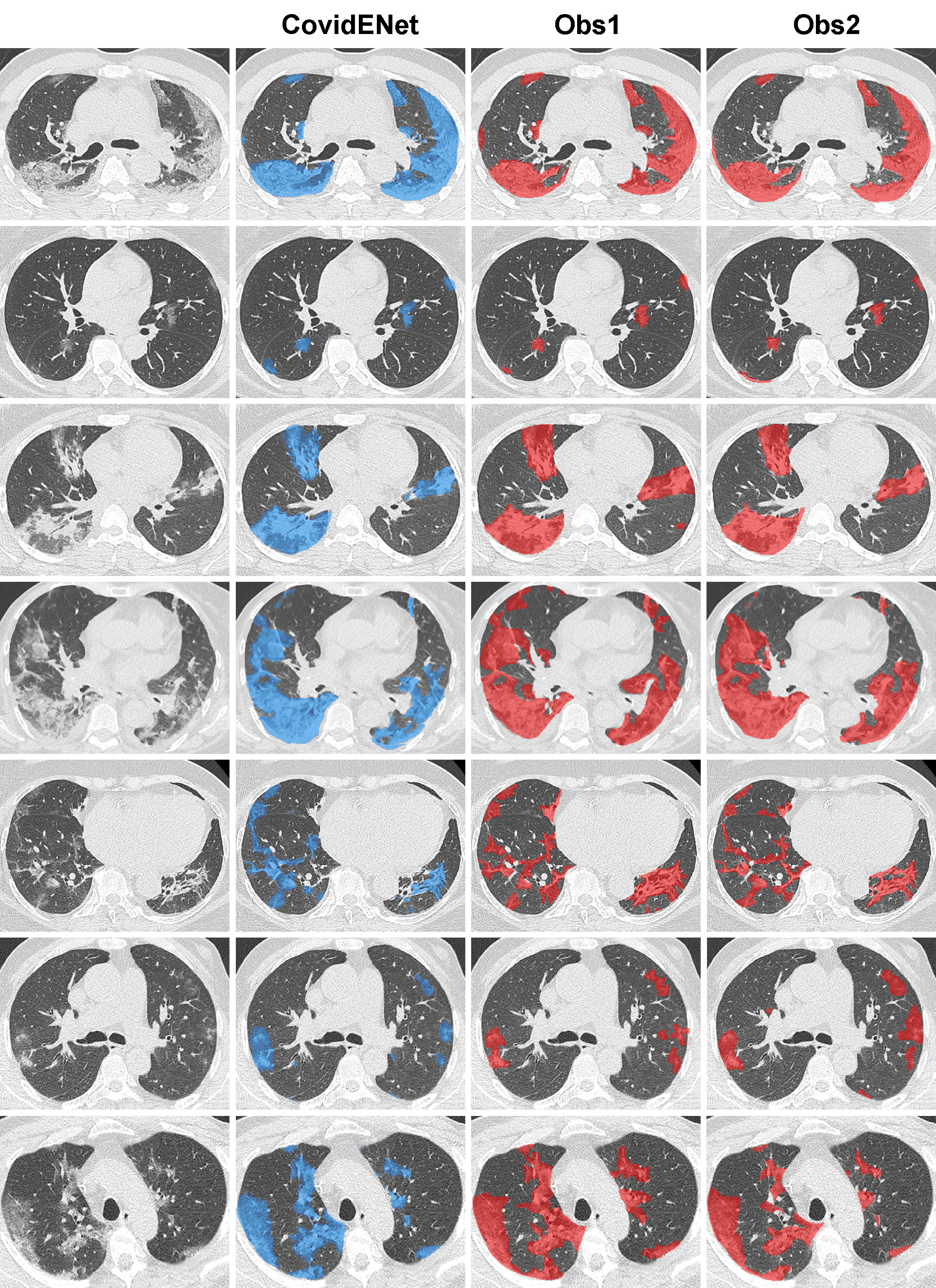}
\caption{Some additional qualitative analysis for the comparison between automated and manual segmentations. Delineation of the diseased areas on chest CT in different slices of COVID-19 patients: From  left to right: Input, AI-segmentation, expert I-segmentation, expert II-segmentation}
\label{fig:exfig3}
\end{figure}

For the 2D experiments we used classic stochastic gradient descent for the optimization with initial learning rate = 0.01, decrease of learning rate = $2.5\cdot10^{-3}$ every $10$ epochs, momentum =$0.9$ and weight decay =$5\cdot10^{-4}$. For the 3D experiments we used the AMSGrad and a learning rate of $0.001$.
The training of a single network for both 2D and 3D network was completed in approximately $12$ hours using a GeForce GTX 1080 GPU, while the prediction for a single CT scan was done in a few seconds. Training and validation curves for one template of AtlasNet and the 3D network are shown in Figure~\ref{fig:fig5}.
Both Dice similarity score and Haussdorff distances were higher with the 2D approach compared to the 3D approach (Figure~\ref{fig:fig2,fig:exfig3}). However, the combination of their probability scores led to a significant improvement. Thus, the ensemble of 2D and 3D architectures was selected for the final COVID-19 segmentation tool.

Moreover, segmentation masks of the lung and heart of all patients were extracted by using ART-Plan software (TheraPanacea, Paris, France). ART-Plan is a CE-marked solution for automatic annotation of organs, harnessing a combination of anatomically preserving and deep learning concepts. This software has been trained using a combination of a transformation and an image loss. The transformation loss penalizes the normalized error between the prediction of the network and the affine registration parameters depicting the registration between the source volume and the whole body scanned. These parameters are determined automatically using a downhill simplex optimization approach. The second loss function of the network involved an image similarity function – the zero-normalized cross correlation loss – that seeks to create an optimal visual correspondence between the observed CT values of the source volume and the corresponding ones at the full body CT reference volume. This network was trained using as input a combination of $360,000$ pairs of CT scans of all anatomies and full body CT scans. These projections used to determine the organs being present on the test volume. Using the transformation between the test volume and the full body CT, we were able to determine a surrounding patch for each organ being present in the volume. These patches were used to train the deep learning model for each full body CT. The next step consisted of creating multiple annotations on the different reference spaces, and for that a 3D fully convolutional architecture was trained for every reference anatomy. This architecture takes as input the annotations for each organ once mapped to the reference anatomy and then seeks to determine for each anatomy a network that can optimally segment the organ of interest similar to the AtlasNet framework used for the disease segmentation. This information was applied for every organ of interest presented in the input CT Scan. In average, $6,600$ samples were used for training per organ after data augmentation. These networks were trained using a conventional dice loss. The final organ segmentation was achieved through a winner takes all approach over an ensemble networks approach. For each organ, and for each full body reference CT a specific network was built, and the segmentation masks generated for each network were mapped back to the original space. The consensus of the recommendations of the different subnetworks was used to determine the optimal label at the voxel level. 

\subsection*{Staging / Prognosis}
As a preprocessing step, all images were resampled by cubic interpolation to obtain isometric voxels with sizes of $1$ mm. Subsequently, disease, lung and heart masks were used to extract $107$ radiomic features~\cite{van2017computational} for each of them (left and right lung were considered separately both for the disease extent and entire lung). The features included first order statistics, shape-based features in 2D and 3D together with texture-based features. Radiomics features were enriched with clinical data available from the image metadata (age, gender), disease extent and number of diseased regions. The minimum and maximum values were calculated for the training and validation cohorts and Min-Max normalization was used to normalize the features, the same values were also applied on the test set.
As a first step, a number of features were selected using a lasso linear model in order to decrease the dimensionality. The lasso estimator seeks to optimize the following objective function:

\begin{equation*}
    \dfrac{||y - Xw||^2_2}{2 n} + \alpha  ||w||_1
\end{equation*}

where $\alpha$ is a constant, $||w||_1$ is the L1-norm of the coefficient vector and n is the number of samples. The Lasso method was used with 200 alphas along a regularization path of length 0.01 and limited to 1000 iterations. The staging/prognosis component was addressed using an ensemble learning approach. First, the training data set was subdivided into training and validation set on the principle of $80\%-20\%$ while respecting that the distribution of classes between the two subsets was identical to the observed one. We have created $10$ subdivisions on this basis and evaluated the average performance of the following supervised classification methods: Nearest Neighbor, \{Linear, Sigmoid, Radial Basis Function, Polynomial Kernel\} Support Vector Machines, Gaussian Process, Decision Trees, Random Forests, AdaBoost, Gaussian Naive Bayes, Bernoulli Naive Bayes, Multi-Layer Perceptron \& Quadratic Discriminant Analysis. Features selection was performed on the training set of each subdivision. In particular, the features selected in at least three subdivision were considered critical and have been used later for the staging and prognosis.
\begin{itemize}
    \item Age
    \item Gender
    \item Disease Extent
    \item From the diseased areas: maximum attenuation, surface, maximum 2D diameter per slice, volume, non-uniformity of the Gray level Size Zone matrix (GLSZM) and non-uniformity of the Gray level Run Length matrix (GLRLM)
    \item From the lung areas: skewness and $90$th percentile
    \item From the area of the heart: non-uniformity on the Gray level Size Zone matrix (GLSZM)
\end{itemize}

These features included first order features (maximum attenuation, skewness and $90$th percentile), shape features (surface, maximum 2D diameter per slice and volume) and texture features (non-uniformity of the GLSZM and GLRLM).

Subsequently, this reduced feature space was considered to be most appropriate for training, and the following $7$ classification methods with acceptable performance, $>60\%$ in terms of balanced accuracy, as well as coherent performance between training and validation, performance decrease $<20\%$ for the balanced accuracy between training and validation, were trained and combined together through a winner takes all approach to determine the optimal outcome (Table~\ref{tab:tab4}). The final selected methods include the \{Linear, Polynomial Kernel, Radial Basis Function\} Support Vector Machines, Decision Trees, Random Forests, AdaBoost, and Gaussian Naive Bayes which were trained and combined together through a winner takes all approach to determine the optimal outcome. To overcome the unbalance of the different classes, each class received a weight inversely proportional to its size. The Support Vector Machines were all three granted a polynomial kernel function of degree $3$ and a penalty parameter of $0.25$. In addition, the one with a Radial Basis Function kernel was granted a kernel coefficient of $3$. The decision tree classifier was limited to a depth of $3$ to avoid overfitting. The random forest classifier was composed of 8 of such trees. AdaBoost classifier was based on a decision tree of maximal depth of $2$ boosted three times. 

\begin{figure}[t!]
\centering
\includegraphics[width=0.9\linewidth]{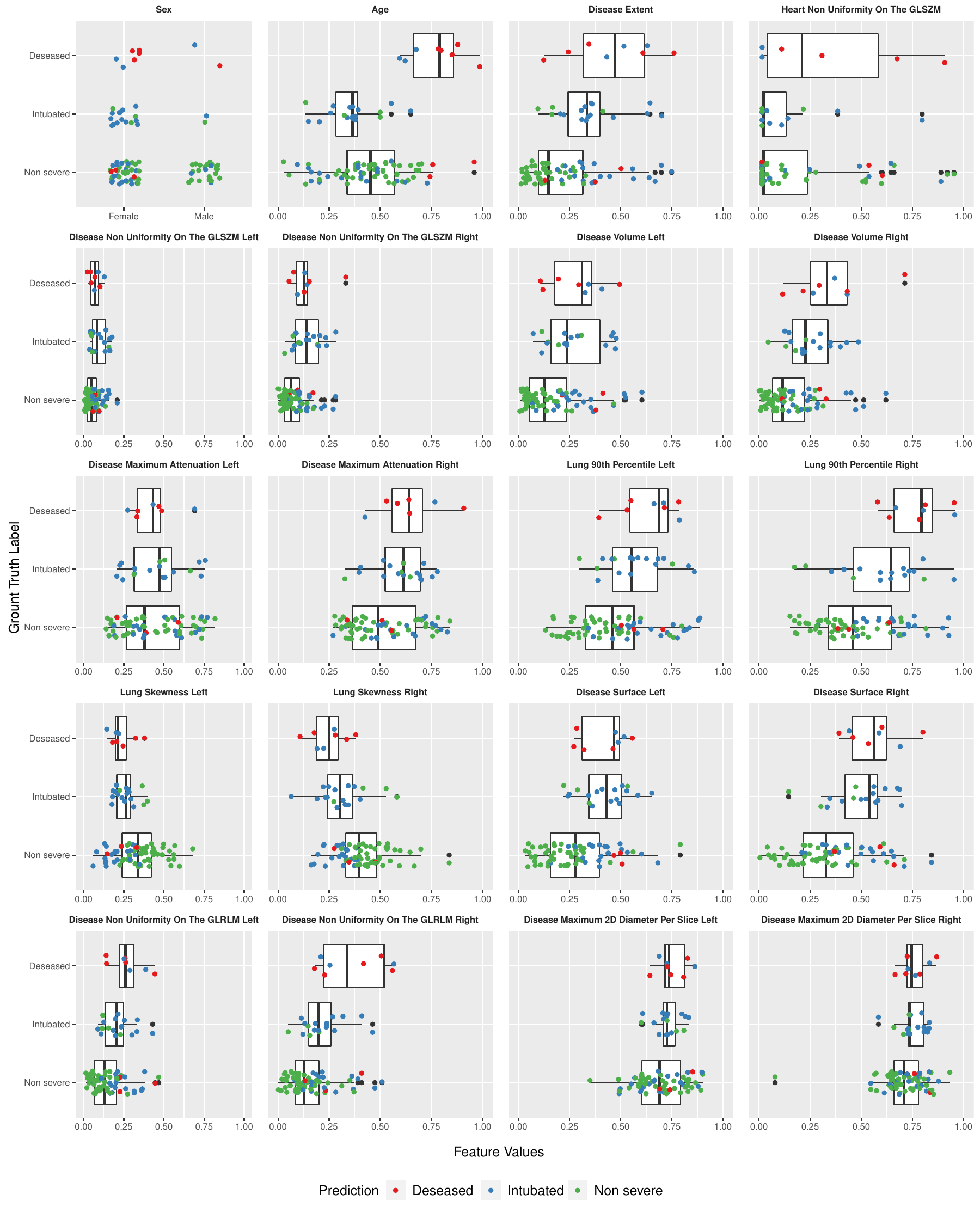}
\caption{Boxplots of the selected features and their association with the predicted outcomes and ground truth labels.}
\label{fig:fig6}
\end{figure}

The classifiers were applied in a hierarchical way, performing first the staging and then the prognosis. More specifically, a majority voting method was applied to classify patients into severe and non-severe cases (Table~\ref{tab:tab6}). Then, another majority voting was applied on the cases predicted as severe only to classify them into intubated or deceased (Table~\ref{tab:tab7}). In such a setup, the correlation of the reported features are summarized in Table~\ref{tab:tab5}. For the hierarchical prognosis on the three classes a voting classifier for the prediction of each class against the others has been applied to aggregate the predicted outcomes from the $7$ selected methods. In the Figure~\ref{fig:fig6} we visualize the distributions of the different features along the ground truth labels and the prediction of the hierarchical classifier for each subject. In particular, all the samples are grouped using their ground truth labels and a boxplot is generated for each group and each feature. Additionally, color coded points are over imposed at each boxplot denoting the prediction label. It is therefore clearly visible that some features such as the disease extent, the age, the shape of the disease and the uniformity seems to be very important on separating the different subjects.

\subsection*{Statistical Analysis}
The statistical analysis for the deep learning-based segmentation framework and the radiomics study was performed using Python $3.7$, Scipy~\cite{virtanen2020scipy}, Scikit-learn~\cite{pedregosa2011scikit}, TensorFlow~\cite{abadi2015tensorflow} and Pyradiomics~\cite{van2017computational} libraries. The dice similarity score (DSC)~\cite{dice1945measures} was calculated to assess the similarity between the $2$ manual segmentations of each CT exam of the test dataset and between manual and automated segmentations. The DSC between manual segmentations served as reference to evaluate the similarity between the automated and the two manual segmentations. Moreover, the Hausdorff distance was also calculated to evaluate the quality of the automated segmentations in a similar manner. Disease extent was calculated by dividing the volume of diseased lung by the lung volume and expressed in percentage of the total lung volume. Disease extent measurement between manual segmentations and between automated and manual segmentations were compared using paired Student's t-tests.

For the stratification of the dataset into the different categories, classic machine learning metrics, namely balanced accuracy, weighted precision, and weighted specificity and sensitivity were used. Moreover, the correlations between each feature and the outcome was computing using a Pearson correlation over the entire dataset.

CT parameters between the $6$ centers were compared using the analysis of variance, while patient characteristics between training/validation and test datasets were compared using chi-square and Student's t-tests.

\begin{table}[h!]
\centering{
\begin{tabular}{|p{20mm}|l|l|l|l|l|l|}
\hline
                          & \multicolumn{3}{|l|}{Training/ validation dataset}                                                                                                                                                                                                           & \multicolumn{3}{|l|}{Testing dataset}                                                                                                                                                                                                                           \\
\cline{2-7}
                          & Center A                                                                        & Center B                                                                            & Center C                                                                           & Center D                                                                           & Center E                                                                            & Center F                                                                           \\
\hline
CT equipment              & Somatom AS+                                                                     & Resolution HD                                                                       & Aquilion Prime                                                                     & Somatom Edge                                                                       & Revolution HD                                                                       & Aquilion Prime                                                                     \\
\hline
Kilovoltage               & 100-120                                                                         & 120                                                                                 & 100-120                                                                            & 100-120                                                                            & 120-140                                                                             & 100-120                                                                            \\
\hline
DLP (mGy.cm)              & \begin{tabular}[c]{@{}l@{}}$109\pm42$\\
{[}44-256{]}\end{tabular}  & \begin{tabular}[c]{@{}l@{}}$306 \pm 104$\\ 
{[}123-648{]}\end{tabular}  & \begin{tabular}[c]{@{}l@{}}$102 \pm 30$\\ 
{[}43-189{]}\end{tabular}   & \begin{tabular}[c]{@{}l@{}}$131 \pm 44$\\ 
{[}55-499{]}\end{tabular}   & \begin{tabular}[c]{@{}l@{}}$177 \pm 48$\\ 
{[}43-276{]}\end{tabular}    & \begin{tabular}[c]{@{}l@{}}$115 \pm 26$\\ {[}75 - 186{]}\end{tabular} \\
\hline
CTDIvol (mGy)             & \begin{tabular}[c]{@{}l@{}}$3.2 \pm1.5$\\ {[}1.2-11.9{]}\end{tabular} & \begin{tabular}[c]{@{}l@{}}$8.7 \pm 2.8$\\ {[}3.9-18.5{]}\end{tabular} & \begin{tabular}[c]{@{}l@{}}$2.7 \pm 0.9$\\ {[}1.0-5.3{]}\end{tabular} & \begin{tabular}[c]{@{}l@{}}$3.2 \pm 0.9$\\ {[}1.4-9.5{]}\end{tabular} & \begin{tabular}[c]{@{}l@{}}$5.5 \pm 1.8$\\ {[}1.2-12.3{]}\end{tabular} & \begin{tabular}[c]{@{}l@{}}$2.5 \pm 0.6$\\ {[}1.7-4.3{]}\end{tabular} \\
\hline
Slice thickness           & 1mm                                                                             & 0.625mm                                                                             & 1mm                                                                                & 0.625mm                                                                            & 1mm                                                                                 & 1mm                                                                                \\
\hline
Convolution Kernel        & i70                                                                             & Lung                                                                                & FC51-FC52                                                                          & i50                                                                                & Lung                                                                                & FC51-FC52                                                                          \\
\hline
Iterative reconstructions & SAFIRE 3                                                                        & ASIR-v 80\%                                                                         & IDR 3D0.67                                                                         & SAFIRE 4                                                                           & ASIR-v 60\%                                                                         & IDR 3D \\
\hline
\end{tabular}
}
\caption{\label{tab:tab1} Acquisition and reconstruction parameters. \textit{Note.— For quantitative variables, data are mean $\pm$ standard deviation, and numbers in brackets are the range. CT = Computed Tomography ; CTDIvol = volume Computed Tomography Dose Index ; DLP = Dose Length Product
* significant difference with p $< 0.001$}}
\end{table}

\begin{table}[t!]
\begin{tabular}{|l|l|l|l|}
\hline
               & \begin{tabular}[c]{@{}l@{}}Training/Validation Dataset\\ (Centers A+B+C; N=50)\end{tabular} & \begin{tabular}[c]{@{}l@{}}Test Dataset\\ (Centers D+E+F; n=130)\end{tabular} & p value \\
\hline
Age (y)        & $57 \pm 17$ {[}26-97{]}                                                        & $59 \pm 16$ {[}17-95{]}                                             & 0.363   \\
\hline
No. of Men     & 31(62)                                                                                      & 87(67)                                                                        & 0.534   \\
\hline
Disease extent* &                                                                                             &                                                                               &         \\
\hline
Manual         & $18.1 \pm$ 14.9 {[}0.3-68.5{]}                                                 & $19.5 \pm 16.5$ {[}1.1-75.7{]}                                   & 0.574   \\
\hline
Automated      & -                                                                                           & $19.9\% \pm 17.7$ {[}0.5-73.2{]}                                  & -       \\
\hline
DLP (mGy.cm)   & $180 \pm 124$ {[}43-527{]}                                                     & $139 \pm 49.0$ {[}43-276{]}                                      & 0.026   \\
\hline
CTDIvol (mGy)  & $4.9 \pm 3.4$ {[}1.0-13.0{]}                                                   & $4.0 \pm 1.9$ {[}1.2-12.3{]}                                     & 0.064  \\
\hline
\end{tabular}
\caption{\label{tab:tab2} Patient characteristics in the datasets used for developing the segmentation tool. \textit{Note. For quantitative variables, data are mean $\pm$ standard deviation, and numbers in brackets are the range. For qualitative variables, data are numbers of patients, and numbers in parentheses are percentages.
CTDIvol = volume Computed Tomography Dose Index; DLP = Dose Length Product
*Percentage of lung volume on CT, calculated on the full volume for the training/validation dataset and 20 slices in the test dataset}}
\end{table}

\begin{table}[t!]
\begin{tabular}{|l|l|l|l|}
\hline
               & \begin{tabular}[c]{@{}l@{}}Training/Validation Dataset\\ (Centers A+B+D*; N=383)\end{tabular} & \begin{tabular}[c]{@{}l@{}}Test Dataset\\ (Centers C+E+F; n=95)\end{tabular} & p value \\
\hline
Age (y)        & $63 \pm 16$ {[}24-98{]}                                                        & $57 \pm 15$ {[}17-97{]}                                             & 0.001   \\
\hline
No. of Men     & 255(67)                                                                                      & 65(68)                                                                        & 0.732   \\
\hline
Disease extent** &  $19.6 \pm 17.0$  {[}0.0-85.1{]} & $22.5 \pm 16.4$  {[}1.1-64.7{]}  & 0.126 \\
\hline
Short-term outcome       &     &       &  \\
\hline
Deceased      & 19(5)& 8(8)&        \\
\hline
Intubated      & 65(17)& 18(19)&        \\
\hline
Alive and Not Intubated & 299(78) & 69(73) &   \\ \hline
DLP (mGy.cm)   & $160 \pm 97$ {[}44-648{]}                                                     & $146 \pm 52.0$ {[} 43-276 {]}                                       & 0.047   \\
\hline
CTDIvol (mGy)  & $4.3 \pm 2.8$ {[}1.2-18.5{]}                                                   & $4.1 \pm 2.0$ {[}1.0-12.3{]}                                     & 0.064  \\
\hline
\end{tabular}
\caption{\label{tab:tab3}  Patient characteristics in the dataset used for the developed prognosis model using radiomics. \textit{Note.— For quantitative variables, data are mean $\pm$ standard deviation, and numbers in brackets are the range. For qualitative variables, data are numbers of patients, and numbers in parentheses are percentages.
CTDIvol = volume Computed Tomography Dose Index; DLP = Dose Length Product, $*$Enlarged by including all eligible patients over the study period, $**$Percentage of lung volume on the whole CT}}
\end{table}

\begin{table}[t!]
\begin{tabular}{|l|l|l|l|l|l|l|l|l|}
\hline
\multirow{2}{*}{Classifier}                                         & \multicolumn{2}{|l|}{\begin{tabular}[c]{@{}l@{}}Balanced Accuracy\end{tabular}}                                                                 & \multicolumn{2}{|l|}{\begin{tabular}[c]{@{}l@{}}Weighted  Precision\end{tabular}}                                                             & \multicolumn{2}{|l|}{\begin{tabular}[c]{@{}l@{}}Weighted Sensitivity\end{tabular}}                                                            & \multicolumn{2}{|l|}{\begin{tabular}[c]{@{}l@{}}Weighted Specificity\end{tabular}}                                                             \\ 
\cline{2-9}
                                                                    & Training                                                                & Validation                                                            & Training                                                              & Validation                                                            & Training                                                              & Validation                                                            & Training                                                              & Validation                                                             \\
\hline
\begin{tabular}[c]{@{}l@{}}Nearest\\ Neighbors\end{tabular}          & \begin{tabular}[c]{@{}l@{}}0.66 \\ $\pm 0.02$\end{tabular} & \begin{tabular}[c]{@{}l@{}}0.55\\ $\pm0.03$\end{tabular} & \begin{tabular}[c]{@{}l@{}}0.88\\ $\pm0.01$\end{tabular} & \begin{tabular}[c]{@{}l@{}}0.73\\ $\pm0.03$\end{tabular} & \begin{tabular}[c]{@{}l@{}}0.85\\ $\pm0.01$\end{tabular} & \begin{tabular}[c]{@{}l@{}}0.78\\ $\pm0.01$\end{tabular} & \begin{tabular}[c]{@{}l@{}}0.47\\ $\pm0.03$\end{tabular} & \begin{tabular}[c]{@{}l@{}}0.32\\ $\pm0.04$\end{tabular}  \\
\hline
\textbf{L-SVM*}                                                     & \begin{tabular}[c]{@{}l@{}}0.71\\ $\pm0.02$\end{tabular}   & \begin{tabular}[c]{@{}l@{}}0.67\\ $\pm0.04$\end{tabular} & \begin{tabular}[c]{@{}l@{}}0.80\\ $\pm0.02$\end{tabular} & \begin{tabular}[c]{@{}l@{}}0.77\\ $\pm0.03$\end{tabular} & \begin{tabular}[c]{@{}l@{}}0.68\\ $\pm0.02$\end{tabular} & \begin{tabular}[c]{@{}l@{}}0.67\\ $\pm0.04$\end{tabular} & \begin{tabular}[c]{@{}l@{}}0.74\\ $\pm0.03$\end{tabular} & \begin{tabular}[c]{@{}l@{}}0.67\\ $\pm0.06$\end{tabular}  \\
\hline
\textbf{P-SVM*}                                                              & \begin{tabular}[c]{@{}l@{}}0.77\\ $\pm0.02$\end{tabular}   & \begin{tabular}[c]{@{}l@{}}0.66\\ $\pm0.04$\end{tabular} & \begin{tabular}[c]{@{}l@{}}0.83\\ $\pm0.01$\end{tabular} & \begin{tabular}[c]{@{}l@{}}0.76\\ $\pm0.03$\end{tabular} & \begin{tabular}[c]{@{}l@{}}0.77\\ $\pm0.03$\end{tabular} & \begin{tabular}[c]{@{}l@{}}0.72\\ $\pm0.03$\end{tabular} & \begin{tabular}[c]{@{}l@{}}0.76\\ $\pm0.03$\end{tabular} & \begin{tabular}[c]{@{}l@{}}0.61\\ $\pm0.07$\end{tabular}  \\
\hline
S-SVM                                                               & \begin{tabular}[c]{@{}l@{}}0.53\\ $\pm0.02$\end{tabular}   & \begin{tabular}[c]{@{}l@{}}0.55\\ $\pm0.05$\end{tabular} & \begin{tabular}[c]{@{}l@{}}0.69\\ $\pm$0.04\end{tabular} & \begin{tabular}[c]{@{}l@{}}0.69\\ $\pm0.04$\end{tabular} & \begin{tabular}[c]{@{}l@{}}0.49\\ $\pm0.08$\end{tabular} & \begin{tabular}[c]{@{}l@{}}0.51\\ $\pm0.1$\end{tabular}  & \begin{tabular}[c]{@{}l@{}}0.57\\ $\pm0.09$\end{tabular} & \begin{tabular}[c]{@{}l@{}}0.58\\ $\pm0.11$\end{tabular}  \\
\hline
\textbf{RBF-SVM*}                                                            & \begin{tabular}[c]{@{}l@{}}0.74\\ $\pm0.02$\end{tabular}   & \begin{tabular}[c]{@{}l@{}}0.67\\ $\pm0.06$\end{tabular} & \begin{tabular}[c]{@{}l@{}}0.82\\ $\pm0.01$\end{tabular} & \begin{tabular}[c]{@{}l@{}}0.77\\ $\pm0.04$\end{tabular} & \begin{tabular}[c]{@{}l@{}}0.66\\ $\pm0.02$\end{tabular} & \begin{tabular}[c]{@{}l@{}}0.63\\ $\pm0.06$\end{tabular} & \begin{tabular}[c]{@{}l@{}}0.81\\ $\pm0.02$\end{tabular} & \begin{tabular}[c]{@{}l@{}}0.71\\ $\pm0.08$\end{tabular} \\
\hline
\begin{tabular}[c]{@{}l@{}}Gaussian\\ 			Process\end{tabular}       & \begin{tabular}[c]{@{}l@{}}0.63\\ 			 $\pm0.03$\end{tabular}                & \begin{tabular}[c]{@{}l@{}}0.59\\ 	$\pm0.04$\end{tabular}               & \begin{tabular}[c]{@{}l@{}}0.82\\ $\pm0.02$\end{tabular}               & \begin{tabular}[c]{@{}l@{}}0.77\\ 	$\pm0.04$\end{tabular}               & \begin{tabular}[c]{@{}l@{}}0.83\\ $\pm0.01$\end{tabular}               & \begin{tabular}[c]{@{}l@{}}0.80\\ 	$\pm0.02$\end{tabular}               & \begin{tabular}[c]{@{}l@{}}0.43\\ 			$\pm0.04$\end{tabular}               & \begin{tabular}[c]{@{}l@{}}0.38\\ $\pm0.05$\end{tabular}                \\
\hline
\begin{tabular}[c]{@{}l@{}}\textbf{Decision}\\ 			\textbf{Tree}*\end{tabular}         & \begin{tabular}[c]{@{}l@{}}0.78\\ 		$\pm0.02$\end{tabular}                 & \begin{tabular}[c]{@{}l@{}}0.68\\ $\pm0.04$\end{tabular}               & \begin{tabular}[c]{@{}l@{}}0.85\\ 	$\pm0.01$\end{tabular}               & \begin{tabular}[c]{@{}l@{}}0.78\\ $\pm0.03$\end{tabular}               & \begin{tabular}[c]{@{}l@{}}0.69\\ $\pm0.03$\end{tabular}               & \begin{tabular}[c]{@{}l@{}}0.64\\ $\pm0.05$\end{tabular}               & \begin{tabular}[c]{@{}l@{}}0.87\\ 			$\pm0.03$\end{tabular}               & \begin{tabular}[c]{@{}l@{}}0.72\\ $\pm0.08$\end{tabular}                \\
\hline
\begin{tabular}[c]{@{}l@{}}\textbf{Random}\\ 			\textbf{Forest*}\end{tabular}         & \begin{tabular}[c]{@{}l@{}}0.78\\ 			$\pm0.02$\end{tabular}                 & \begin{tabular}[c]{@{}l@{}}0.64\\ 	$\pm0.06$\end{tabular}               & \begin{tabular}[c]{@{}l@{}}0.84\\ $\pm0.02$\end{tabular}               & \begin{tabular}[c]{@{}l@{}}0.75\\ 		$\pm0.04$\end{tabular}               & \begin{tabular}[c]{@{}l@{}}0.74\\ 	$\pm0.03$\end{tabular}               & \begin{tabular}[c]{@{}l@{}}0.65\\ 	$\pm0.07$\end{tabular}               & \begin{tabular}[c]{@{}l@{}}0.82\\ 		$\pm0.04$\end{tabular}               & \begin{tabular}[c]{@{}l@{}}0.63\\ 	$\pm0.09$\end{tabular}                \\
\hline
\begin{tabular}[c]{@{}l@{}}Multi-Layer\\ 			Perceptron\end{tabular} & \begin{tabular}[c]{@{}l@{}}0.83\\ 			$\pm0.04$\end{tabular}                 & \begin{tabular}[c]{@{}l@{}}0.58\\ 			$\pm0.03$\end{tabular}               & \begin{tabular}[c]{@{}l@{}}0.90\\ 			$\pm0.02$\end{tabular}               & \begin{tabular}[c]{@{}l@{}}0.72\\ 			$\pm0.02$\end{tabular}               & \begin{tabular}[c]{@{}l@{}}0.9\\ 			$\pm0.02$\end{tabular}                & \begin{tabular}[c]{@{}l@{}}0.73\\ 			$\pm0.03$\end{tabular}               & \begin{tabular}[c]{@{}l@{}}0.77\\ 			$\pm0.06$\end{tabular}               & \begin{tabular}[c]{@{}l@{}}0.43\\ 			$\pm0.05$\end{tabular}                \\
\hline
\textbf{AdaBoost*}                                                           & \begin{tabular}[c]{@{}l@{}}0.8\\ 			 $\pm0.03$\end{tabular}                 & \begin{tabular}[c]{@{}l@{}}0.63\\ 			$\pm0.03$\end{tabular}               & \begin{tabular}[c]{@{}l@{}}0.86\\ 			$\pm0.01$\end{tabular}               & \begin{tabular}[c]{@{}l@{}}0.75\\ 			$\pm0.03$\end{tabular}               & \begin{tabular}[c]{@{}l@{}}0.76\\ 			$\pm0.07$\end{tabular}               & \begin{tabular}[c]{@{}l@{}}0.66\\ 			$\pm0.07$\end{tabular}               & \begin{tabular}[c]{@{}l@{}}0.83\\ 			$\pm0.08$\end{tabular}               & \begin{tabular}[c]{@{}l@{}}0.6\\ 			$\pm0.11$\end{tabular}                 \\
\hline
\begin{tabular}[c]{@{}l@{}}\textbf{Gaussian}\\ 			\textbf{Naive Bayes*}\end{tabular}  & \begin{tabular}[c]{@{}l@{}}0.66\\ 			 $\pm0.02$\end{tabular}                & \begin{tabular}[c]{@{}l@{}}0.63\\ $\pm0.05$\end{tabular}   & \begin{tabular}[c]{@{}l@{}}0.76\\ 	$\pm0.01$\end{tabular}               & \begin{tabular}[c]{@{}l@{}}0.74\\ 			$\pm0.03$\end{tabular}               & \begin{tabular}[c]{@{}l@{}}0.74\\ 			$\pm0.01$\end{tabular}               & \begin{tabular}[c]{@{}l@{}}0.72\\ 			$\pm0.04$\end{tabular}               & \begin{tabular}[c]{@{}l@{}}0.57\\ 			$\pm0.03$\end{tabular}               & \begin{tabular}[c]{@{}l@{}}0.53\\ 			$\pm0.07$\end{tabular}                \\
\hline
\begin{tabular}[c]{@{}l@{}}Bernouilli\\ 			Naive Bayes\end{tabular} & \begin{tabular}[c]{@{}l@{}}0.51\\ 			$\pm0.01$\end{tabular}                 & \begin{tabular}[c]{@{}l@{}}0.50	\\ $\pm0.01$\end{tabular}   & \begin{tabular}[c]{@{}l@{}}0.74\\ 			$\pm0.05$\end{tabular}               & \begin{tabular}[c]{@{}l@{}}0.63\\ 			$\pm0.05$\end{tabular}               & \begin{tabular}[c]{@{}l@{}}0.78\\ 			$\pm0.0$\end{tabular}                & \begin{tabular}[c]{@{}l@{}}0.77\\ 			$\pm0.01$\end{tabular}               & \begin{tabular}[c]{@{}l@{}}0.24\\ 			$\pm0.01$\end{tabular}               & \begin{tabular}[c]{@{}l@{}}0.22\\ 			$\pm0.02$\end{tabular}                \\
\hline
QDA                                                                 & \begin{tabular}[c]{@{}l@{}}0.72\\ 			 $\pm0.01$\end{tabular}                & \begin{tabular}[c]{@{}l@{}}0.60\\ 			$\pm0.04$\end{tabular}   & \begin{tabular}[c]{@{}l@{}}0.82\\ 			$\pm0.01$\end{tabular}               & \begin{tabular}[c]{@{}l@{}}0.73\\ 			$\pm0.03$\end{tabular}               & \begin{tabular}[c]{@{}l@{}}0.82\\ 			$\pm0.01$\end{tabular}               & \begin{tabular}[c]{@{}l@{}}0.75\\ 			$\pm0.03$\end{tabular}               & \begin{tabular}[c]{@{}l@{}}0.62\\ 			$\pm0.02$\end{tabular}               & \begin{tabular}[c]{@{}l@{}}0.45\\ 			$\pm0.07$\end{tabular}  \\
\hline
\end{tabular}
\caption{\label{tab:tab4} Performances of the 13 evaluated classifiers. \textit{Note.— Asterixis indicates the 7 classifiers reporting a balanced accuracy higher than 0.60 and that were finally selected,
L-SVM = Support Vector Machine with a linear kernel; L-SVM = Support Vector Machine with a polynomial kernel, L-SVM = Support Vector Machine with a sigmoid kernel, RBF-SVM = Support Vector Machine with a Radial Basis Function, QDA = Quadratic Discriminant Analysis}}
\end{table}

\begin{table}[t!]
\begin{tabular}{|l|l|l|l|l|l|l|l|}
\hline
\multicolumn{2}{|l|}{\multirow{2}{*}{Features}}                                                        & \multicolumn{6}{|l|}{Correlation}                                                                                                    \\
\cline{3-8}
\multicolumn{2}{|l|}{}                                                                                 & \multicolumn{2}{|l|}{Severe vs Non Severe} & \multicolumn{2}{|l|}{Intubated vs Diseased} & \multicolumn{2}{|l|}{Hierarchical Classifier} \\
\hline
\multicolumn{2}{|l|}{Age}                                                                              & \multicolumn{2}{|l|}{-0.0681}              & \multicolumn{2}{|l|}{0.6745}                & \multicolumn{2}{|l|}{0.1853}                  \\
\hline
\multicolumn{2}{|l|}{Gender}                                                                           & \multicolumn{2}{|l|}{-0.1517}              & \multicolumn{2}{|l|}{0.0211}                & \multicolumn{2}{|l|}{0.1443}                  \\
\hline
\multicolumn{2}{|l|}{Disease Extent}                                                                   & \multicolumn{2}{|l|}{-0.3809}              & \multicolumn{2}{|l|}{-0.0280}               & \multicolumn{2}{|l|}{0.3491}                  \\
\hline
Heart                    & \begin{tabular}[c]{@{}l@{}}non-uniformity \\ on the GLSZM\end{tabular}    & \multicolumn{2}{|l|}{-0.1953}              & \multicolumn{2}{|l|}{-0.0935}               & \multicolumn{2}{|l|}{0.1537}                  \\
\cline{2-8}
                         &                                                                           & Left                & Right              & Left                & Right               & Left                 & Right                \\
\hline
\multirow{2}{*}{Lung}    & Skewness                                                                  & 0.3324              & 0.3675             & 0.1018              & -0.0229             & -0.2967              & -0.3447              \\
\cline{2-8}
                         & 90th Percentile                                                           & -0.2808             & -0.3221            & -0.0576             & 0.0035              & 0.2526               & 0.3119               \\
\hline
\multirow{6}{*}{Disease} & \begin{tabular}[c]{@{}l@{}}Maximum\\ attenuation\end{tabular}             & -0.0867             & -0.1978            & 0.0885              & 0.0035              & 0.0958               & 0.1846               \\
\cline{2-8}
                         & Surface                                                                   & -0.3382             & -0.3289            & -0.0174             & 0.0381              & 0.3034               & 0.3112               \\
\cline{2-8}
                         & \begin{tabular}[c]{@{}l@{}}Maximum 2D\\ diameter per\\ slice\end{tabular} & -0.2304             & -0.2236            & 0.1238              & 0.1010              & 0.2288               & 0.2183               \\
\cline{2-8}
                         & Volume                                                                    & -0.3592             & -0.3971            & -0.0489             & 0.0840              & 0.3241               & 0.3863               \\
\cline{2-8}
                         & \begin{tabular}[c]{@{}l@{}}Non-uniformity \\ \\ on the GLSZM\end{tabular} & -0.2944             & -0.3448            & -0.1379             & -0.0805             & 0.2375               & 0.3004               \\
\cline{2-8}
                         & \begin{tabular}[c]{@{}l@{}}Non-uniformity\\ on the GLRLM\end{tabular}     & -0.3280             & -0.3372            & -0.0219             & 0.0829              & 0.3000               & 0.3319  \\            
\hline
\end{tabular}
\caption{\label{tab:tab5} Correlation between outcome and the 12 selected features. \textit{Note.—Note.— GLSZM  =Gray level Size Zone matrix , GLRLM = Gray level Run Length matrix}}
\end{table}
\begin{table}[b!]
\begin{tabular}{|l|l|l|l|l|l|l|l|l|}
\hline
Classifier                                                     & \multicolumn{2}{l|}{\begin{tabular}[c]{@{}l@{}}Balanced\\ Accuracy\end{tabular}} & \multicolumn{2}{l|}{\begin{tabular}[c]{@{}l@{}}Weighted\\ Precision\end{tabular}} & \multicolumn{2}{l|}{\begin{tabular}[c]{@{}l@{}}Weighted\\ Sensitivity\end{tabular}} & \multicolumn{2}{l|}{\begin{tabular}[c]{@{}l@{}}Weighted\\ Specificity\end{tabular}} \\
\cline{2-9}
                                                               & Training                                 & Test                                 & Training                                  & Test                                 & Training                                   & Test                                  & Training                                   & Test                                  \\
\hline
L-SVM                                                          & 0.71                                     & 0.7                                  & 0.8                                       & 0.76                                 & 0.68                                       & 0.67                                  & 0.75                                       & 0.73                                  \\
\hline
P-SVM                                                          & 0.76                                     & 0.67                                 & 0.82                                      & 0.73                                 & 0.75                                       & 0.69                                  & 0.76                                       & 0.65                                  \\
\hline
RBF-SVM                                                        & 0.72                                     & 0.71                                 & 0.81                                      & 0.79                                 & 0.65                                       & 0.63                                  & 0.79                                       & 0.79                                  \\
\hline
Decision Tree                                                  & 0.76                                     & 0.65                                 & 0.85                                      & 0.74                                 & 0.65                                       & 0.56                                  & 0.88                                       & 0.74                                  \\
\hline
Random Forest                                                  & 0.76                                     & 0.7                                  & 0.82                                      & 0.76                                 & 0.71                                       & 0.65                                  & 0.8                                        & 0.75                                  \\
\hline
AdaBoost                                                       & 0.77                                     & 0.67                                 & 0.83                                      & 0.74                                 & 0.71                                       & 0.62                                  & 0.82                                       & 0.71                                  \\
\hline
\begin{tabular}[c]{@{}l@{}}Gaussian\\ Naive Bayes\end{tabular} & 0.65                                     & 0.73                                 & 0.76                                      & 0.77                                 & 0.73                                       & 0.76                                  & 0.57                                       & 0.69                                  \\
\hline
\begin{tabular}[c]{@{}l@{}}\textbf{Ensemble}\\ \textbf{Classifier}\end{tabular}  & 0.75                                     & 0.74                                 & 0.82                                      & 0.79                                 & 0.69                                       & 0.69                                  & 0.81                                       & 0.79      \\
\hline
\end{tabular}
\caption{\label{tab:tab6}Performances of each of the 7 individual classifiers and of the ensemble classifier to differentiate between patient with severe and non-severe short-term outcome. \textit{Note.— L-SVM = Support Vector Machine with a linear kernel; L-SVM = Support Vector Machine with a polynomial kernel, RBF-SVM = Support Vector Machine with a Radial Basis Function kernel}}
\end{table}

\begin{table}[t!]
\begin{tabular}{|l|l|l|l|l|l|l|l|l|}
\hline
Classifier                                                     & \multicolumn{2}{l|}{\begin{tabular}[c]{@{}l@{}}Balanced\\ Accuracy\end{tabular}} & \multicolumn{2}{l|}{\begin{tabular}[c]{@{}l@{}}Weighted\\ Precision\end{tabular}} & \multicolumn{2}{l|}{\begin{tabular}[c]{@{}l@{}}Weighted\\ Sensitivity\end{tabular}} & \multicolumn{2}{l|}{\begin{tabular}[c]{@{}l@{}}Weighted\\ Specificity\end{tabular}} \\
\cline{2-9}
                                                               & Training                                 & Test                                 & Training                                  & Test                                 & Training                                   & Test                                  & Training                                   & Test                                  \\
\hline
L-SVM                                                          & 0.84                                     & 0.78                                 & 0.87                                      & 0.84                                 & 0.81                                       & 0.85                                  & 0.87                                       & 0.72                                  \\
\hline
P-SVM                                                          & 0.95                                     & 0.81                                 & 0.96                                      & 0.9                                  & 0.95                                       & 0.88                                  & 0.95                                       & 0.74                                  \\
\hline
RBF-SVM                                                        & 0.9                                      & 0.62                                 & 0.92                                      & 0.83                                 & 0.90                                       & 0.77                                  & 0.9                                        & 0.48                                  \\
\hline
Decision Tree                                                  & 0.97                                     & 0.75                                 & 0.98                                      & 0.87                                 & 0.98                                       & 0.85                                  & 0.96                                       & 0.65                                  \\
\hline
Random Forest                                                  & 0.94                                     & 0.75                                 & 0.96                                      & 0.87                                 & 0.96                                       & 0.85                                  & 0.92                                       & 0.65                                  \\
\hline
AdaBoost                                                       & 1                                        & 0.62                                 & 1                                         & 0.83                                 & 1                                          & 0.77                                  & 1                                          & 0.48                                  \\
\hline
\begin{tabular}[c]{@{}l@{}}Gaussian\\ Naive Bayes\end{tabular} & 0.82                                     & 0.75                                 & 0.86                                      & 0.87                                 & 0.83                                       & 0.85                                  & 0.8                                        & 0.65                                  \\
\hline
\begin{tabular}[c]{@{}l@{}} \textbf{Ensemble}\\ \textbf{Classifier}\end{tabular}  & 0.96                                     & 0.81                                 & 0.97                                      & 0.9                                  & 0.96                                       & 0.88                                  & 0.95    & 0.74 \\ \hline                                 
\end{tabular}
\caption{\label{tab:tab7} Performances of each of the 7 individual classifiers and of the ensemble classifier to differentiate between intubated and deceased patients. \textit{Note.— L-SVM = Support Vector Machine with a linear kernel; L-SVM = Support Vector Machine with a polynomial kernel, RBF-SVM = Support Vector Machine with a Radial Basis Function kernel}}
\end{table}
\bibliography{sample}







\end{document}